\documentclass[conference]{IEEEtran}

\RequirePackage{amsmath,amssymb,amsfonts}
\RequirePackage{algorithmic}
\RequirePackage{graphicx}
\RequirePackage{textcomp}
\RequirePackage{xcolor}
\RequirePackage{float}
\RequirePackage[style=numeric,maxbibnames=10,sorting=none]{biblatex}
\RequirePackage[colorlinks, citecolor = red]{hyperref}
\RequirePackage[ruled]{algorithm2e}
\RequirePackage{booktabs}
\RequirePackage{setspace}
\RequirePackage{smartdiagram}
\def\BibTeX{{\rm B\kern-.05em{\sc i\kern-.025em b}\kern-.08em
    T\kern-.1667em\lower.7ex\hbox{E}\kern-.125emX}}
 
\RequirePackage{array}
\newcolumntype{L}[1]{>{\raggedright\arraybackslash}p{#1}}
\newcolumntype{C}[1]{>{\centering\arraybackslash}p{#1}}
\newcolumntype{R}[1]{>{\raggedleft\arraybackslash}p{#1}}

\usepackage{color}

\begin{document}

\title{Semi-Automated Knowledge Engineering and Process Mapping for Total Airport Management\\
}

\author{
\IEEEauthorblockN{Darryl Teo, Adharsha Sam, Chuan Shen Marcus Koh, Rakesh Nagi, Nuno Antunes Ribeiro}
\IEEEauthorblockA{Aviation Studies Institute}
Singapore University of Technology and Design\\
Singapore \\
\{darryl\_teo \textbar \ adharsha\_devahi \textbar \ marcus\_koh \textbar \ rakesh\_nagi \textbar \ nuno\_ribeiro\}@sutd.edu.sg
}

\maketitle

\begin{abstract}
    Documentation of airport operations is inherently complex due to extensive technical terminology, rigorous regulations, proprietary regional information, and fragmented communication across multiple stakeholders. The resulting data silos and semantic inconsistencies present a significant impediment to the Total Airport Management (TAM) initiative. This paper presents a methodological framework for constructing a domain-grounded, machine-readable Knowledge Graph (KG) through a dual-stage fusion of symbolic Knowledge Engineering (KE) and generative Large Language Models (LLMs).
    
    The framework employs a scaffolded fusion strategy in which expert-curated KE structures guide LLM prompts to facilitate the discovery of semantically aligned knowledge triples. We evaluate this methodology on the Google LangExtract library and investigate the impact of context window utilization by comparing localized segment-based inference with document-level processing. Contrary to prior empirical observations of long-context degradation in LLMs, document-level processing improves the recovery of non-linear procedural dependencies.
    
    To ensure the high-fidelity provenance required in airport operations, the  proposed framework fuses a probabilistic model for discovery and a deterministic algorithm for anchoring every extraction to its ground source. This ensures absolute traceability and verifiability, bridging the gap between "black-box" generative outputs and the transparency required for operational tooling. Finally, we introduce an automated framework that operationalizes this pipeline to synthesize complex operational workflows from unstructured textual corpora.
\end{abstract}


\section{Introduction}
Organizations now gather vast volumes of data from a wide range of sources, with individual departments and business units often managing their own datasets. However, many barriers to information sharing exist, including conflicts of interest, regulations \cite{Kim2016} and low interoperability standards \cite{Saberi2025}. As such, this information often ends up locked within separate systems, ranging from spreadsheets to databases to technical reports. This insulation of data obscures necessary information despite it having already been acquired by another entity within the same business unit. Furthermore, while technical jargon facilitates clarity within a given occupation, it becomes a fertile ground for miscommunication when members of different operations use different terminology to refer to the same object \cite{Fiset2023}. Such inconsistencies can have severe consequences. A primary cause of the tragic Tenerife Airport Disaster in 1977 was a breakdown in communication attributable to non-standard terminology; poor visibility and the pilot's misinterpretation of phrases such as "OK" and "takeoff" as clearance instructions ultimately led to the catastrophic collision of two B747 aircraft and 583 fatalities \cite{Weick1990}.

The scale of contemporary data infrastructures necessitates automated integration frameworks to mitigate the risks associated with organizational data silos. In this context, establishing data provenance is paramount, providing a formal trace of the data origin and computational processes governing its ingestion \cite{Buneman2001}. This work investigates the utility of LLMs in facilitating domain-specific knowledge fusion for the construction of logically consistent KGs. In this architecture, an ontology serves as the formal, machine-readable specification of a shared conceptualization, defining the classes, properties, and constraints of the domain, while the KG populates this framework with interconnected entities and relationships extracted from operational data. By grounding LLM extractions in the nuances of airport operations through high-quality few-shot prompting, and aligning the KG structure with the NASA Air Traffic Management (ATM) ontology, the framework ensures semantic alignment and rigorous data provenance across complex, multi-stakeholder procedural workflows \cite{NASA2017}.

This paper presents a systematic evaluation of the LangExtract framework \cite{Goel2025} for structured information extraction, specifically examining the impact of varying  scales on the aircraft turn-round process from the EUROCONTROL Airport Collaborative Decision-Making (A-CDM) manual \cite{EUROCONTROL2017}.The A-CDM framework is an industry standard designed to improve operational efficiency by sharing real-time data between pilots, ATCs, and ground handlers through a sequence of 16 Milestones that track a flight's progress and A-CDM Platform updates. This document serves as a robust benchmark due to its description of a complex, time-sensitive workflow. Its logical triggers and state transitions provide a rigorous environment for evaluating the framework's ability to extract precise procedural knowledge. Establishing this KE framework supports the TAM initiative, a EUROCONTROL concept that seeks to move toward proactive, performance-driven operations. In particular, this work addresses the need for a holistic view of airport processes, providing the high-fidelity data integration necessary to synchronize airside and landside performance into a single, predictive source of truth.

The selection of LangExtract is predicated on its capacity for: (i) high-fidelity source grounding, ensuring data provenance; (ii) optimized extraction architectures for long-context windows; and (iii) deterministic structured output generation. These capabilities are particularly salient for automatically synthesizing machine-readable KGs from unstructured operational documentation. Through empirical evaluation on the A-CDM milestones, we show that LangExtract can reliably extract procedural entities and relationships under both page-level and document-level inference settings. Contrary to prior studies, document-level processing improves the recovery of non-linear procedural dependencies while preserving sentence-level provenance for all extracted knowledge triples.

Subsequently, we evaluate the utility of the resulting KG. By anchoring analyses in canonical KG concepts, the architecture enables automated generation of artifacts including process maps, simulations, and scenario analyses. In pursuit of aviation requirements for interpretability, provenance, and stakeholder attribution, the methodology defines an automated swimlane generation framework with source grounding (implementation available on GitHub\footnote{\url{https://github.com/darrylteo/autoKE}}). The principal contribution of this work is the systematic integration and evaluation of algorithmic toolchains for synthesizing operational KGs and analytical artifacts in alignment with the TAM initiative \cite{EUROCONTROL2025}.

\textit{Structure.} The remainder of this paper is organized as follows: \S \ref{sec:lit} provides a comprehensive review of the relevant literature. Section \ref{sec:method} details the methodological framework, including prompt engineering, KG modeling principles, the LangExtract-based extraction pipeline, and the algorithmic generation of swimlane diagrams. Section \ref{sec:res} presents the experimental setup and an empirical evaluation of the results. Section \ref{sec:future} discusses subsequent applications of domain-grounded KGs. Finally, Section \ref{sec:conc} offers concluding remarks and outlines directions for future research.

\section{Literature Review}
\label{sec:lit}
The evolution of robust semantic modeling frameworks exists at the intersection of formal knowledge representation and automated information extraction. While LLMs have transformed the paradigm of information retrieval through high-dimensional vector representations, maintaining the deterministic rigor required for safety-critical domains necessitates a synergy with established KE algorithmic fusion frameworks \cite{BarShalom2011}. This section evaluates the transition from manually-curated ontological schemas to adaptive, LLM-driven architectures, identifying the technical trade-offs between formal consistency and extraction scalability.

Subsection \ref{sub:ke} examines foundational KE methodologies, focusing on the utility of formal structures, specifically KGs, in ensuring interoperability and situational awareness across heterogeneous data sources. While traditional KE relies on expert-led curation to ensure semantic accuracy, it is limited by the scalability bottlenecks of manual annotation. We therefore explore the role of LLMs and their tuning mechanisms including prompt engineering and context window optimization as critical control mechanisms. These emerging techniques serve to tune the probabilistic outputs of LLMs to adhere to the high standards of formal logical schema, thereby ensuring the accuracy and verifiability of the extracted data.

Subsection \ref{sub:keavi} contextualizes these methodologies within the aviation domain, tracing the progression from foundational frameworks, such as the NASA ATM ontology \cite{NASA2017}, to contemporary LLM-integrated systems. By reviewing these developments, we establish the current boundaries of KE its applications in safety-critical operational modeling.

\subsection{Knowledge Engineering}
\label{sub:ke}
Domain-specific KE remains bottlenecked by manual expert curation to resolve complex semantic relationships \cite{anantharangachar2013ontology, wong2022domain}. Despite the integration of domain-optimized BERT variants, high-density technical jargon often limits performance, with F1 scores frequently plateauing near 70\% \cite{durmaz2024ontology}. While semi-automated NLP pipelines and alignment strategies attempt to bridge this gap, they scale poorly due to significant "cold start" dependencies and the administrative requirement of extensive manual labeling \cite{amdouni2025semi}. Consequently, there is a distinct lack of fully autonomous frameworks capable of synthesizing formal KGs without extensive expert intervention. 

Recent advances leverage LLMs to automate KG generation from unstructured documents \cite{Zhang2024, Kommineni2024, Incitti2024}. While effective on small datasets, these models struggle to scale to the extensive corpora typical of real-world applications and often necessitate human-in-the-loop validation. Although prompt engineering enhances output accuracy \cite{Pan2023, Jiang2019, Ji2022}, the probabilistic nature of LLMs remains a challenge. However, incorporating formal logical structures, such as using existing KGs to constrain prompt architecture, has yielded significant improvements \cite{Pan2023, Incitti2024, Chen2021}; for instance, KnowPrompt \cite{Chen2021} reported accuracy gains from 39.9\% to 53.1\%.

Efforts to improve scalability and accuracy have focused on tuning and refinement for long context windows, which frequently suffer from the "lost-in-the-middle" phenomenon \cite{An2024}. Mitigation strategies include chunking, parallel processing, and iterative extraction \cite{Goel2025}, alongside algorithmic developments such as self-extension, positional interpolation, short-length recovery, and Recurrent Context Compression \cite{Jin2024, Ding2024, Huang2024}. Despite these advances, such approaches often encounter prohibitive computational costs, marginal performance gains, or reconstruction errors due to excessive instruction length. Critically, no existing fully automated pipeline performs extraction, KG construction, and process map generation without extensive expert guidance or tedious annotation.

\subsection{Aviation Applications of Knowledge Engineering} 
\label{sub:keavi}
Historically, KE in aviation has been confined to aircraft design and manufacturing within vertically integrated corporate structures \cite{Stjepandic2015, Dadzie2008}. While these frameworks successfully leveraged semantic search to optimize internal manufacturing pipelines \cite{LaRocca2009, Emberey2007}, they rely on centralized, proprietary data repositories. A critical research gap remains in the automated synthesis of KGs for the broader airport ecosystem—a fragmented domain defined by decentralized interactions among independent stakeholders with disparate operational protocols and regulatory constraints.

Within the aviation literature, the NASA ATM ontology \cite{NASA2017} represents a foundational effort in domain formalization, providing comprehensive taxonomies for ATM and trajectory logic. However, while it effectively models flight-centric sequences, such as takeoff and landing, its scope does not fully encapsulate the heterogeneous, multi-stakeholder ground operations (e.g., baggage handling or refueling) critical to TAM. This creates a procedural void in modeling the high-density, collaborative workflows required for integrated operations. Consequently, existing frameworks lack the granular, cross-stakeholder interoperability necessary for a unified semantic representation of the airport ecosystem.

More recently, LLMs have been introduced to parse the vast volumes of unstructured technical aviation documents. State-of-the-art implementations, including AviationGPT \cite{wang2024aviationgpt} and AviationCopilot \cite{zhang2026aviationcopilot}, demonstrate the efficacy of domain-specific LLMs in providing conversational assistance to pilots and engineers. These tools excel at natural-language querying across extensive document collections, frequently reporting high accuracy (exceeding 90\%) through fine-tuning on domain-specific corpora. However, because these models function primarily as black-box algorithms, data provenance remains elusive; the path from the source text to generated response is often obscured by the model's probabilistic nature. Consequently, a disparity remains between the utility of conversational LLMs and the formal requirements of airport operational modeling, where absolute verifiability and grounding in authoritative sources are non-negotiable.

\subsection{Research Gap and Motivation}
Despite advances in knowledge-driven systems, the automated synthesis of verifiable operational models from unstructured text remains an open challenge in multi-stakeholder, safety-critical environments. Airport domain documentation is characterised by dense technical nomenclature, heterogeneous abbreviations, and tightly coupled procedural dependencies. Under these conditions, general-purpose LLMs can achieve high recall yet consistently fail to satisfy the deterministic, auditable requirements of operational decision support. Existing ontological efforts, such as the NASA ATM ontology \cite{NASA2017}, supply rigorous terminological scaffolding but lack the assertional and procedural density required to represent ground operations across independent stakeholders. Conversely, recent domain-adapted LLM deployments, including AviationGPT \cite{wang2024aviationgpt} and AviationCopilot \cite{zhang2026aviationcopilot}, demonstrate strong conversational and retrieval performance; however, their black-box architectures fundamentally preclude traceability and authoritative source grounding, attributes that are non-negotiable in safety-critical operational contexts.

A secondary and largely unaddressed gap concerns the downstream translation of extracted knowledge into actionable operational artifacts. Current pipelines treat KG construction as a terminal objective, providing no principled mechanism for converting structured graph representations into human-interpretable process visualizations. This limitation is particularly acute in multi-stakeholder environments where technical jargon and operational workflows diverge across organizational boundaries, and where explicit stakeholder attribution is a prerequisite for regulatory compliance and coordination.

This work addresses both gaps through a systematic evaluation of the LangExtract framework as an engine for domain-specific knowledge fusion. We assess the effect of varying context window length on extraction veracity by contrasting segment-level inference against document-level inference, thereby quantifying the empirical cost of the "lost-in-the-middle" phenomenon in a structured extraction setting. LangExtract's architecture is further evaluated on its provision of explicit source text grounding, a capability routinely absent from general-purpose models, ensuring that every extracted assertion remains anchored to a verifiable textual provenance.

Building on the resulting KG, we introduce an algorithmic pipeline for the automated generation of swimlane diagrams. By deriving process visualizations directly from the formal graph structure, the framework ensures that each procedural step is both machine-interpretable and attributed to a responsible stakeholder entity. This dual grounding in both formal semantics and explicit organizational accountability supports compliance with stringent safety regulations and advances the interoperability objectives of the TAM initiative \cite{EUROCONTROL2025}.

\section{Methodology}
\label{sec:method}
\begin{figure}[ht]
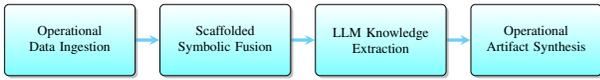

\centering
\resizebox{0.45\textwidth}{!}{
\smartdiagramset{
    border color=black,
    set color list={white!50!cyan,white!50!cyan,white!50!cyan,white!50!cyan},
    back arrow disabled=true,
    font=\large,
    text width=3.8cm,
    module minimum width=3.5cm,
    module minimum height=2.2cm,
    module x sep=4.8
}
\smartdiagram[flow diagram:horizontal]{
    Operational Data Ingestion,
    Scaffolded Symbolic Fusion, 
    LLM Knowledge Extraction, 
    Operational Artifact Synthesis
}}
\caption{Knowledge Engineering Architecture for A-CDM Milestones}
\label{fig:archi}
\end{figure}

The proposed architecture implements a scaffolded symbolic fusion strategy (see Fig.\ \ref{fig:archi}) designed to bridge the gap between deterministic domain knowledge and probabilistic generative models. The pipeline begins with Operational Data Ingestion, which focuses on the textual corpora of the 16 A-CDM Milestones. This phase incorporates data cleaning and normalization (as necessary) to resolve jargon inconsistencies, ensuring the reliability of the downstream extraction process. While this implementation centers on text documents, the architecture is designed to accommodate future expansion into multimodal inputs, specifically the conversion of audio and visual operational recordings into LLM-readable data formats.

The core of the methodology lies in the Scaffolded Symbolic Fusion stage. Here, an expert-curated KG provides the structural constraints used to generate high-fidelity prompts and few-shot examples for the LLM Triple Extraction phase. By using symbolic logic to scaffold the generative model, the system ensures that the extracted entities and relationships are semantically aligned with established aviation standards. This dual-stage process culminates in Operational Artifact Synthesis, where the newly populated KG is algorithmically converted into stakeholder-attributed process maps and swimlane diagrams, providing a verifiable and unified source of truth for airport management.

This remainder of this section outlines the architectural framework of the proposed system. Subsection \ref{sub:kg} formalizes the KG representation and the foundational KGs that govern our semantic schema. Subsequently, \S \ref{sub:le} details the implementation of LangExtract, formalizing our approach to prompt engineering and context window optimization. Finally, we define the evaluation metrics employed to quantify extraction veracity and structural consistency.

\subsection{Knowledge Graphs}
\label{sub:kg}
A KG is an explicit, machine-readable model of the domain: a controlled vocabulary of classes, properties, individuals, and the relationships that connect them. Serving as a single authoritative source of truth, KGs facilitate data association across heterogeneous datasets \cite{Xu2025, Pan2023}, support richer queries and analytics, and enable reasoning — for example, inferring that a "jet bridge" and a "remote stand" are mutually disjoint entities. By standardizing terminology, the KG acts as a shared semantic layer that harmonizes data interpretation across disparate systems. This stands in contrast to traditional documentation tools such as process maps and commercial workflow software, which typically lack standardized vocabularies, produce siloed representations, and cannot be consumed by external applications. In safety-critical, multi-stakeholder environments, the formal logical consistency and computational reusability of a KG are therefore not merely convenient properties but operational requirements.

Transitioning from system modeling to the extraction of authoritative stakeholder knowledge presents significant operational and technical challenges. In the aviation sector, business units often operate under strict privacy constraints, effectively acting as internal competitors who guard proprietary operational logic, metrics or other datasets. As a result, knowledge elicitation becomes fragmented, where definitions vary, access to examples is limited, and essential details may be withheld. Beyond organizational barriers, the high operational cost and time requirements of traditional knowledge elicitation, such as stakeholder workshops and interviews, further limit the scalability of this approach. These constraints make it hard to understand the domain, identify canonical concepts, or reconcile semantic differences across teams. 

To mitigate these constraints, we adopt a modular disclosure strategy drawn from the NASA ontology \cite{NASA2017}, with the KG architected as layers, and are conditionally disseminated depending on the stakeholder. The base layer, accessible to all parties, contains only high-level classes and process structures that do not expose proprietary detail, and additional modules are shared selectively as required. Where complete information remains unavailable, simplifying assumptions are introduced, for example, treating certain processes as black boxes with defined inputs and outputs, or modeling competing units as separate subdomains that interact only through governed interfaces. Leveraging the capacity of KGs for incremental fulfillment \cite{Villela2005}, this approach ensures the model remains functional under data access constraints while allowing progressive refinement as model maturity increases.

\begin{figure}[ht]
    \centering
    \includegraphics[width=0.4\linewidth]{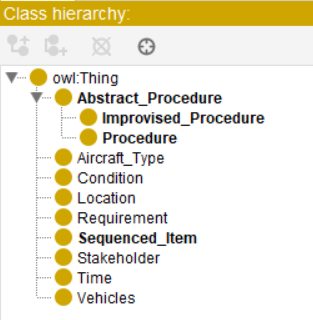}
    \caption{Protégé Airport Operations Class Schema}
    \label{fig:protegeclasses}
\end{figure}

The proposed schema is scaffolded upon the NASA ATM ontology \cite{NASA2017} and the Changi Airport Petri maps \cite{Sng2019}, and is implemented in Protégé, a widely adopted open-source ontology editor. This schema serves as a formalized template, which may be populated manually, or via structured extraction pipelines as described in \S \ref{sec:method}. While the full class hierarchy encompasses both landside and airside operations (see Fig.\ \ref{fig:protegeclasses}), this work focuses on three classes: Procedure, Sequenced\_Item, and Stakeholder; and two object properties: hasNext and hasStakeholder. The broader class structure is retained to ensure extensibility and alignment with existing domain KGs \cite{NASA2017, Sng2019}. 

\subsection{LangExtract}
\label{sub:le}
The transition from unstructured technical corpora to a formal KG requires an extraction engine that accurately captures entities and their relationships. Standard LLM-based extraction is susceptible to hallucination, where the model synthesizes plausible but non-existent information \cite{Ji2022}. This problem is particularly acute in aviation, where safety-critical documentation demands absolute data provenance. The capacity to trace every extracted entity or relationship back to its exact source, down to the specific paragraph and document, is therefore a functional necessity.

A primary objective of this work is to evaluate model scalability and mitigate the performance degradation typically observed as the context window expands. While modern LLMs claim support for windows exceeding 128k tokens, prior empirical evidence suggests a catastrophic spike in perplexity once sequences surpass native training limits \cite{Ding2024}, often manifesting as structural incoherence or nonsensical output. We address these potential limitations through the LangExtract framework \cite{Goel2025}, which enforces structural determinism via schema-guided extraction that maintains explicit data provenance through traceable references. By constraining both the prompt architecture and the model's response to the formal classes defined in curated few-shot examples, LangExtract ensures that the output is immediately machine-readable and compatible with the KG construction pipeline without post-extraction cleaning.

\begin{figure}[ht]
    \centering
    \includegraphics[width=0.9\linewidth]{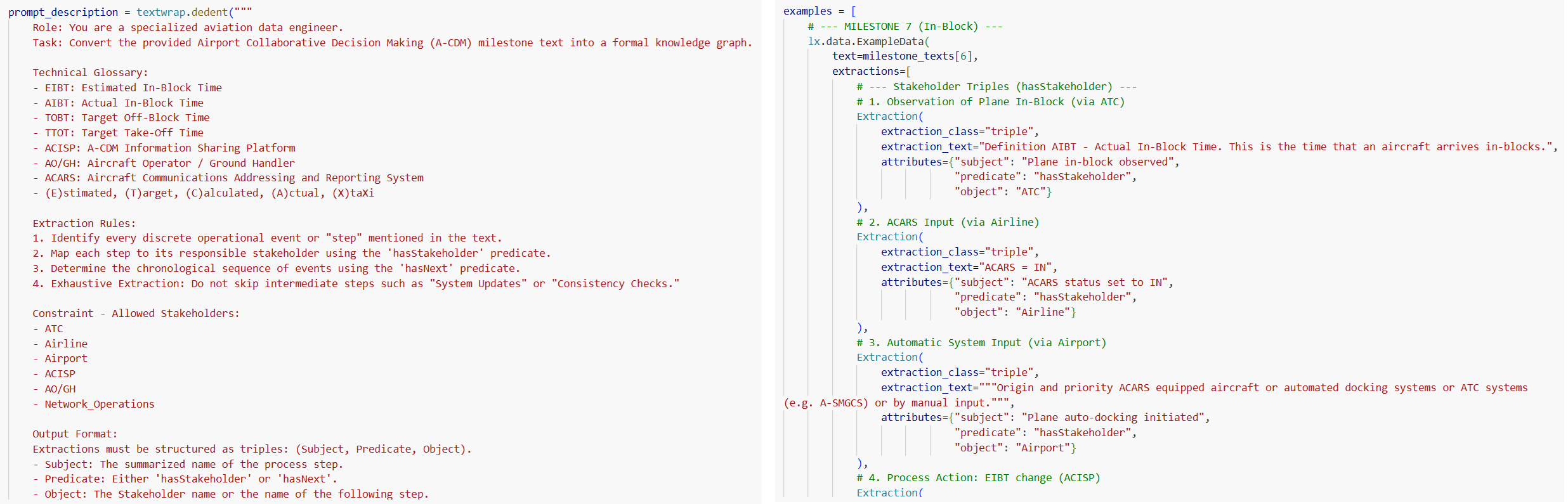}
    \caption{LangExtract Prompt and Few-shot Examples}
    \label{fig:leprompt}
\end{figure}

As illustrated in Fig. \ref{fig:leprompt}, both the few-shot exemplars and the system prompts are explicitly governed by the manually curated Knowledge Graph (KG). These components integrate the class hierarchies and relational schemata defined in \S \ref{sub:kg} with a targeted corpus of examples from a single A-CDM Milestone \cite{EUROCONTROL2017}. The primary function of this initial ontological seed is to facilitate scaffolded symbolic fusion. By providing a formal structural baseline, it complements the LangExtract framework, ensuring logical alignment across all few-shot prompts and grounding the model’s probabilistic extractions within the deterministic constraints of the airport operations domain.

The extraction performance is measured by Recall ($R$), Precision ($P$), and the F1-Score ($F1$), defined as:$$R = \frac{TP}{TP+FN}, \quad P = \frac{TP}{TP+FP}, \quad F1 = 2 \cdot \frac{P \cdot R}{P+R}$$ Recall indicates the framework's sensitivity in identifying all mandatory operational procedures, Precision measures the structural integrity of extracted triples against the domain schema, and the F1-Score represents the overall reliability of the information retrieval process.

Central to this evaluation is the effect of context window length on semantic integrity. Given the perplexity cliff observed in technical corpora exceeding native transformer limits \cite{Ding2024}, we process a 16-page operational manual (approx. 10k tokens) under two modalities: page-level inference to maximize attention density and mitigate the "lost-in-the-middle" phenomenon \cite{An2024}, and document-level inference to evaluate framework stability over extended, high-density sequences. Data provenance is assessed by auditing the sentence-level source references generated by LangExtract, ensuring that every triple in the resulting KG satisfies the traceability requirements necessary for deployment in safety-critical environments.

\subsection{Swimlane Diagrams}
\label{sub:swim}
Beyond static representation, the proposed framework enables dynamic operational modeling in multi-stakeholder environments. By serving as a unified semantic ground truth, the KG mitigates coordination failures arising from conflicting technical interpretations \cite{Fiset2023}. The architecture transforms machine-readable procedural relationships into structured operational artifacts, such as swimlane diagrams, ensuring every step is explicitly attributed to a stakeholder in alignment with the interoperability objectives of the TAM initiative \cite{EUROCONTROL2025}.

\begin{figure}[ht]
    \centering
    \includegraphics[width=0.9\linewidth]{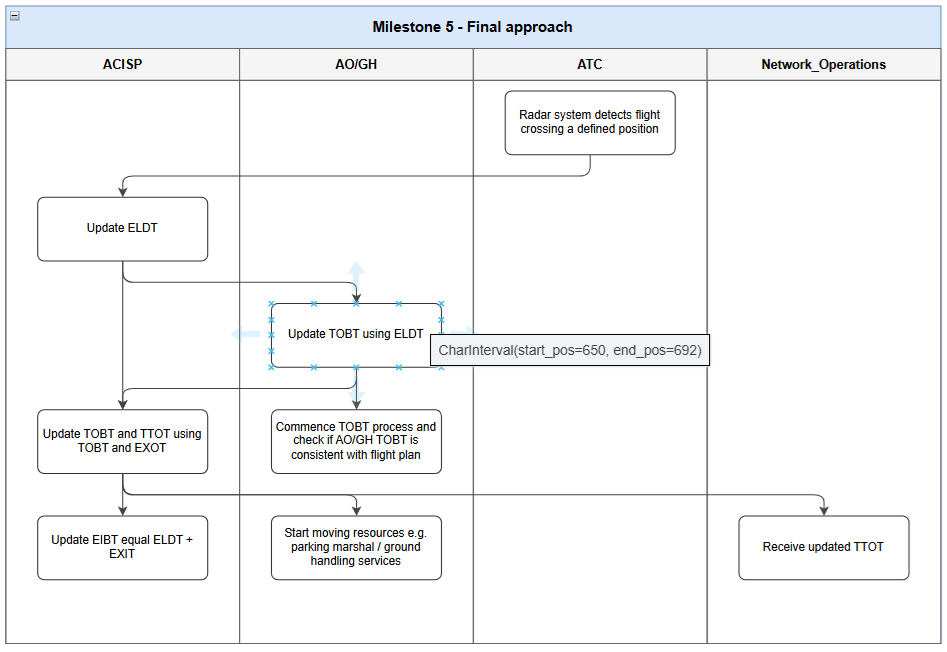}
    \caption{Swimlane Diagram for the Airport Collaborative Decision-Making Process}
    \label{fig:swimlane}
\end{figure}

Utilizing the formalized schema, entities, and relationships from the A-CDM KG \cite{EUROCONTROL2017}, an automated pipeline was developed for generating swimlane diagrams. These diagrams partition processes into parallel lanes per stakeholder, making ownership explicit across airlines, ground handlers, and air traffic services. Such visualization is well-suited for operator training and for surfacing responsibility gaps during process design or incident review (see Fig. \ref{fig:swimlane}). The transformation from KGs to swimlane diagrams is formalized in Algorithm \ref{alg:drawswimlane}. This procedure employs a modified topological sort using Breadth-First Search (BFS) to resolve procedural dependencies, computing the longest path to each node for vertical positioning, while horizontal placement is assigned by mapping each process to its respective stakeholder lane.

\begin{algorithm}
\caption{\texttt{drawSwimlane($V, E, S$)}}
\label{alg:drawswimlane}
\setstretch{0.9}
\tcp{Get vertex depth using BFS}
\For{vertex $v \in V$}{
    \If{$inDegree[v] == 0$}{
        $depth[v] = 1$\\
        $Queue.enqueue(v)$
    }
}

\While{$Queue$ is not empty}{
    $v = Queue.dequeue()$\\    
    \For{each neighbor $u$ of $v$}{
        $depth[u] = \max(depth[u], depth[v] + 1)$\\
        $inDegree[u] = inDegree[u] - 1$\\
        \If{$inDegree[u] == 0$}{
            $Queue.enqueue(u)$
        }
    }
}

\tcp{Plot on stakeholder and depth}
\For{stakeholder $s \in S$}{
    \texttt{drawLane(s)}\\
    \For{each $v \in V$ where $s_v = s$}{
        \texttt{drawNode(depth[v], s)}
    }
}

\For{edge $e \in E$}{
    \texttt{drawArrow($e_{start}, e_{end}$)}
}
\end{algorithm}

To satisfy the verifiability requirements of safety-critical applications, data provenance is integrated directly into the diagram. Each node rendered by \texttt{drawNode()} is linked to its sentence-level source coordinates extracted by LangExtract, anchoring the swimlane in authoritative text. This verifiable audit trail mitigates the risks of LLM hallucination and fosters trust in automated knowledge engineering pipelines.

\section{Experimental Results}
\label{sec:res}

This section evaluates LangExtract’s performance in synthesizing a coherent KG from the A-CDM manual \cite{EUROCONTROL2017} across varying context window lengths. Milestone processes were first converted from PDF to plain text to be read into LangExtract \cite{pdf242026}. To benchmark the system and enable knowledge fusion, we manually curated a ground-truth KG that exhaustively encodes sequential relations (\texttt{hasNext}) and stakeholder responsibilities (\texttt{hasStakeholder}). We ensured deterministic evaluation through a high-fidelity few-shot prompt (Fig. \ref{fig:leprompt}) derived from the ground truth. This prompt facilitates scaffolded symbolic fusion by constraining the LLM’s probabilistic outputs to the formal domain schema \cite{Incitti2024}.

\begin{figure}[ht]
    \centering
    \includegraphics[width=0.9\linewidth]{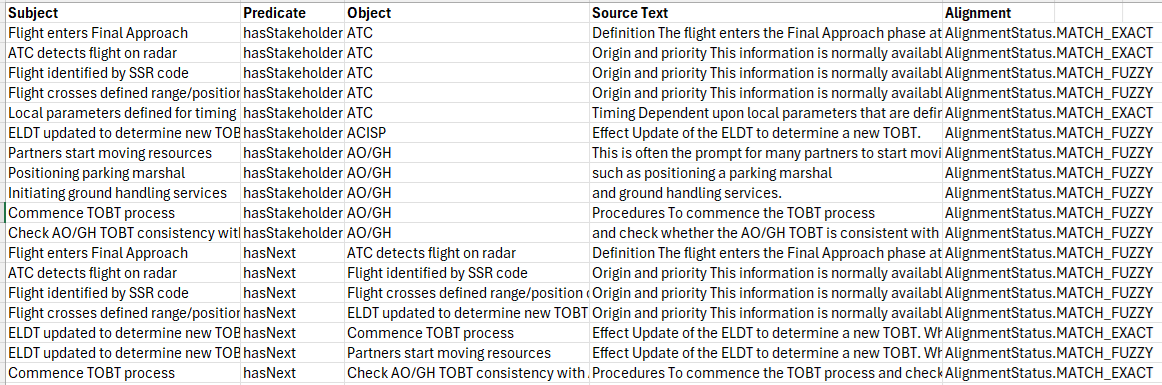}
    \caption{LangExtract Structured Extraction Output With Source Reference}
    \label{fig:leop}
\end{figure}

As illustrated in Fig.\ \ref{fig:leop}, LangExtract enforces a structured output schema and preserves the provenance of each triple. This deterministic structure enables seamless conversion of extracted entities into KGs via the \texttt{owlready2} library \cite{JeanBaptiste2021}, bypassing the noise typical of unstructured NLP pipelines.

To address the complexity of the extracted procedural knowledge, we performed an exhaustive manual evaluation against the curated ground-truth KG and source A-CDM document. Each extracted triple was classified as a True Positive (TP) or False Positive (FP), while omitted triples were marked as False Negatives (FN), with sentence-level provenance verified for every entry to ensure high-confidence traceability. Extraction performance is quantified using the metrics defined in \S \ref{sub:le}, where alignment between the ground-truth KG and extracted triples are defined as: a \textbf{TP} is a correctly retrieved triple, an \textbf{FN} is an omitted triple from the curated KG, and an \textbf{FP} is an erroneous retrieval absent from the ground truth.

Comparative analysis using two context window configurations: short-context inference, which restricted processing to single-page chunks to maximize LLM attention density, and long-context inference, which processed the entire 16-page corpus in a single pass to evaluate framework robustness against the "lost-in-the-middle" phenomenon \cite{An2024}. All experiments utilized a fixed configuration (\texttt{max\_workers=1}) leveraging \texttt{gemini-2.5-flash} with all remaining parameters at default values to ensure reproducibility.

\begin{table}[htbp]
  \centering
  \caption{Performance Information Retrieval Metrics of LangExtract on the EUROCONTROL A-CDM Manual}
    \begin{tabular}{lrr}
    \toprule
        & \multicolumn{1}{l}{Short-Context} & \multicolumn{1}{l}{Long-Context} \\
    \midrule \midrule
    TP    & 440   & 442 \\
    FP    & 18    & 15 \\
    FN    & 13    & 8 \\
    Precision (\textit{P}) & 0.961 & 0.967 \\
    Recall (\textit{R}) & 0.971 & 0.982 \\
    F1 Score (\textit{F1}) & 0.966 & 0.975 \\
    \bottomrule
    \end{tabular}%
  \label{tab:res1}%
\end{table}%

The main results are given in Table \ref{tab:res1}. LangExtract attains high extraction fidelity across both configurations: short-context inference yields \textit{P} = 0.961, \textit{R} = 0.971, and \textit{F1} = 0.966, while long-context inference yields \textit{P} = 0.967, \textit{R} = 0.982, and \textit{F1} = 0.975. Document-level processing reduced FNs by 38.5\% (13 to 8), produced a modest reduction in FPs, and improved \textit{R} and \textit{F1} by 1.1 and 0.9 percentage points respectively.

Notably, long-context outperformed short-context inference, contrary to the theoretical expectation of performance degradation under extended sequences. This improvement is attributed to the spatial locality of the documented airport processes: as most related terms and procedural dependencies occur within a three-page span, the expanded context window allows LangExtract to maintain coherence across process boundaries. By avoiding the information loss associated with arbitrary text fragmentation, the model successfully minimized triple omissions and achieved measurably higher recall.

The primary driver of FNs across both configurations is the inversion of logical flow, where the effect of a procedure is frequently described prior to the procedure itself. This semantic inversion confuses the LLM under limited context, as insufficient surrounding text prevents correct reordering of events. The long-context window mitigates this by providing the broader context required to identify these inversions and correctly assign temporal ordering between cause and effect.

LangExtract classifies string alignment between source text and extracted triples into three categories: MATCH\_EXACT (verbatim extraction), MATCH\_FUZZY (semantic mapping with minor variation), and MATCH\_LESSER (partial mapping with significant deviation). Results by alignment level are detailed in Table \ref{tab:res2}.

Comparing MATCH\_EXACT and MATCH\_FUZZY, the relative occurrence of FPs is higher for MATCH\_FUZZY under both configurations, consistent with the expectation that triples lacking a precise verbatim match are more susceptible to hallucination. MATCH\_LESSER alignments produced no FPs in either configuration, likely due to the low frequency of this class and the model requiring high confidence to assign a partial mapping, thereby suppressing false generations.

\begin{table}[htbp]
  \centering
  \caption{Provenance Information Retrieval Metrics of LangExtract on the EUROCONTROL A-CDM Manual}
    \begin{tabular}  {p{2.7cm} R{1cm} R{1cm} R{1cm} R{1cm}}
    \toprule
          & \multicolumn{2}{c}{Short-Context} & \multicolumn{2}{c}{Long-Context} \\ \cmidrule(lr){2-3} \cmidrule(lr){4-5}
          & \multicolumn{1}{r}{FP} & \multicolumn{1}{r}{TP} & \multicolumn{1}{r}{FP} & \multicolumn{1}{r}{TP} \\
    \midrule \midrule
    MATCH\_EXACT & 4     & 138   & 3     & 149 \\
    MATCH\_FUZZY & 14    & 273   & 12    & 276 \\
    MATCH\_LESSER & 0     & 22    & 0     & 17 \\
    \bottomrule
    \end{tabular}%
  \label{tab:res2}%
\end{table}%

We evaluated provenance accuracy by defining a successful match as a triple that corresponds verbatim to or is directly inferable from the source text. LangExtract successfully mapped all extracted triples to their respective source sentences, providing a verifiable audit trail. This reliability stems from LangExtract's native fusion of the LLM’s probabilistic outputs with deterministic string alignment methods, specifically \texttt{text.find()} and \texttt{difflib.SequenceMatcher()}, ensuring robust traceability across the entire KG.

Empirical results demonstrate that LangExtract effectively bridges unstructured aviation documentation and structured, verifiable KGs. Contrary to theoretical expectations, long-context processing yielded superior performance, with the expanded context window reducing FNs caused by logical inversions in the source text. Robust provenance mapping further ensures that every extracted triple is traceable to its source, fulfilling a critical transparency requirement for deployment in safety-critical environments.

\section{Future Work}
\label{sec:future}
\begin{figure}[ht]
    \centering
    \includegraphics[width=0.87\linewidth]{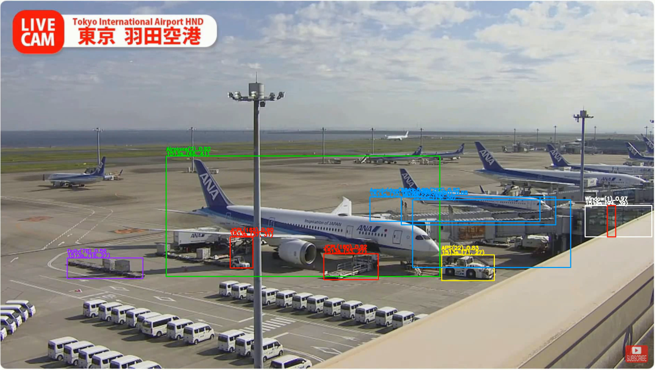}
    \caption{Sample video analytics of Haneda Airport Operations}
    \label{fig:haneda_sample}
\end{figure}

Airport ground operations are subject to improvisations that frequently bypass standard procedural documentation, making full-scale analysis of efficiency impacts, safety risks, and stakeholder accountability a significant challenge. To address this, we propose a real-time operations verification framework that fuses multimodal data, including video analytics (see Fig.\ \ref{fig:haneda_sample}) and transponder signals, against canonical sequences defined in the KG, enabling automatic flagging of procedural deviations and initiation of recovery protocols. A critical challenge in this context is entity resolution and data association \cite{Madani2019, Hogan2012}; specifically, mapping low-level sensor detections to high-level KG entities. We intend to explore Named Entity Recognition techniques and Hidden Markov Models to bridge this semantic gap in future work.

Beyond real-time monitoring, recurring deviations and novel operational patterns detected through transponder and audiovisual evidence can inform refinements to both documented procedures and the KG itself, creating a feedback loop where live operational data continuously improves the accuracy, completeness, and adaptability of the airport's knowledge base, in direct alignment with the goals of the TAM initiative \cite{EUROCONTROL2025}.

\section{Conclusion}
\label{sec:conc}
This paper has presented a methodological framework for the semi-automated construction of domain-grounded knowledge graphs, addressing the persistent challenges of data silos and semantic inconsistency across airport operations. At the core of this framework is a scaffolded symbolic fusion strategy, which uses expert-curated ontological structures to constrain and direct the generative capabilities of LLMs. Systematic evaluation of this approach via the LangExtract library demonstrated high-fidelity information extraction from unstructured technical corpora, with both long and short context configurations yielding precision, recall, and F1-scores exceeding 95\%. While segment-based inference optimizes attention density for entity recognition, document-level processing remains indispensable for capturing global procedural dependencies in workflows where logical flows are non-linear or inverted. The fusion of deterministic string alignment methods, specifically \texttt{text.find()} and \texttt{difflib.SequenceMatcher()}, with the probabilistic outputs of LangExtract provided complete source provenance across all extracted triples, reinforcing the traceability and reliability of the synthesized semantic data.

The integration of these triples into a validated KG establishes an interoperable, machine-readable foundation for the TAM initiative \cite{EUROCONTROL2025}, enabling the automated generation of stakeholder-attributed swimlane diagrams that render operational processes both computationally interpretable and directly actionable. Unlike conversational LLM-based systems \cite{wang2024aviationgpt, zhang2026aviationcopilot}, the proposed approach preserves explicit links to authoritative source text, providing a verifiable audit trail for each extracted relationship, a prerequisite for deploying LLM-derived knowledge in safety-critical aviation environments.

Looking forward, the framework constitutes a scalable foundation for real-time operational verification through multimodal data fusion, wherein formal KG logic is combined with sensor inputs such as video analytics and transponder data to support proactive detection of procedural deviations. Collectively, this work advances the goals of modern airport management by reducing the opacity of generative AI outputs through constrained extraction and deterministic provenance, offering a practical and auditable pathway for integrating LLMs into complex, multi-stakeholder operational workflows. 

\printbibliography

@article{amdouni2025semi,
  title={Semi-Automatic Building of Ontologies from Unstructured French Texts: Industrial Case Study: E. Amdouni et al.},
  author={Amdouni, Emna and Belfadel, Abdelhadi and Gagnant, Maxence and Renault, Isabelle and Kierszbaum, Samuel and Carrion, Jeremy and Dussartre, Matthieu and Tmar, Sana},
  journal={Data Science and Engineering},
  pages={1--23},
  year={2025},
  publisher={Springer}
}

@article{anantharangachar2013ontology,
  title={Ontology guided information extraction from unstructured text},
  author={Anantharangachar, Raghu and Ramani, Srinivasan and Rajagopalan, S},
  journal={arXiv preprint arXiv:1302.1335},
  year={2013}
}

@InProceedings{An2024,
  author     = {An, Shengnan and Chen, Weizhu and Lin, Zeqi and Lou, Jian-Guang and Ma, Zexiong and Zheng, Nanning},
  booktitle  = {Advances in Neural Information Processing Systems 37},
  title      = {Make Your LLM Fully Utilize the Context},
  year       = {2024},
  pages      = {62160--62188},
  publisher  = {Neural Information Processing Systems Foundation, Inc. (NeurIPS)},
  series     = {NeurIPS 2024},
  collection = {NeurIPS 2024},
  doi        = {10.52202/079017-1986},
}

@Book{BarShalom2011,
  author    = {Bar-Shalom, Yaakov},
  editor    = {Peter K. Willett and Xin Tian},
  publisher = {YBS Publishing},
  title     = {Tracking and data fusion},
  year      = {2011},
  address   = {Storrs, CT},
  isbn      = {9780964831278},
  pagetotal = {1235},
  ppn_gvk   = {668210931},
  subtitle  = {A handbook of algorithms},
}

@InBook{Buneman2001,
  author    = {Buneman, Peter and Khanna, Sanjeev and Wang-Chiew, Tan},
  pages     = {316--330},
  publisher = {Springer Berlin Heidelberg},
  title     = {Why and Where: A Characterization of Data Provenance},
  year      = {2001},
  isbn      = {9783540445036},
  booktitle = {Database Theory — ICDT 2001},
  doi       = {10.1007/3-540-44503-x_20},
  issn      = {0302-9743},
}

@Article{Chen2021,
  author    = {Chen, Xiang and Zhang, Ningyu and Xie, Xin and Deng, Shumin and Yao, Yunzhi and Tan, Chuanqi and Huang, Fei and Si, Luo and Chen, Huajun},
  title     = {KnowPrompt: Knowledge-aware Prompt-tuning with Synergistic Optimization for Relation Extraction},
  year      = {2021},
  copyright = {arXiv.org perpetual, non-exclusive license},
  doi       = {10.48550/ARXIV.2104.07650},
  publisher = {arXiv},
}

@Article{Dadzie2008,
  author    = {Dadzie, A.-S. and Bhagdev, R. and Chakravarthy, A. and Chapman, S. and Iria, J. and Lanfranchi, V. and Magalhães, J. and Petrelli, D. and Ciravegna, F.},
  journal   = {Journal of Intelligent Manufacturing},
  title     = {Applying semantic web technologies to knowledge sharing in aerospace engineering},
  year      = {2008},
  issn      = {1572-8145},
  month     = jun,
  number    = {5},
  pages     = {611--623},
  volume    = {20},
  doi       = {10.1007/s10845-008-0141-1},
  publisher = {Springer Science and Business Media LLC},
}

@Misc{Ding2024,
  author    = {Ding, Yiran and Zhang, Li Lyna and Zhang, Chengruidong and Xu, Yuanyuan and Shang, Ning and Xu, Jiahang and Yang, Fan and Yang, Mao},
  title     = {LongRoPE: Extending LLM Context Window Beyond 2 Million Tokens},
  year      = {2024},
  copyright = {Creative Commons Attribution Non Commercial No Derivatives 4.0 International},
  doi       = {10.48550/ARXIV.2402.13753},
  publisher = {arXiv},
}

@article{durmaz2024ontology,
  title={An ontology-based text mining dataset for extraction of process-structure-property entities},
  author={Durmaz, Ali Riza and Thomas, Akhil and Mishra, Lokesh and Murthy, Rachana Niranjan and Straub, Thomas},
  journal={Scientific data},
  volume={11},
  number={1},
  pages={1112},
  year={2024},
  publisher={Nature Publishing Group UK London}
}

@InProceedings{Emberey2007,
  author    = {Emberey, C.L. and Milton, N.R. and Berends, J.P.T.J. and Van Tooren, M.J.L. and Van Der Elst, S.W.G. and Vermeulen, B.},
  booktitle = {7th AIAA ATIO Conf, 2nd CEIAT Int’l Conf on Innov and Integr in Aero Sciences,17th LTA Systems Tech Conf; followed by 2nd TEOS Forum},
  title     = {Application of Knowledge Engineering Methodologies to Support Engineering Design Application Development in Aerospace},
  year      = {2007},
  month     = sep,
  publisher = {American Institute of Aeronautics and Astronautics},
  doi       = {10.2514/6.2007-7708},
}

@TechReport{EUROCONTROL2017,
  author = {EUROCONTROL},
  title  = {Airport CDM Implementation},
  year   = {2017},
}

@Article{EUROCONTROL2025,
  author = {EUROCONTROL},
  title  = {Total airport management: Driving the integration of airports into the ATM network by sharing information in a timely manner.},
  year   = {2025},
  url    = {https://www.eurocontrol.int/project/total-airport-management},
}

@Article{Fiset2023,
  author    = {Fiset, John and Bhave, Devasheesh P. and Jha, Nilotpal},
  journal   = {Journal of Management},
  title     = {The Effects of Language-Related Misunderstanding at Work},
  year      = {2023},
  issn      = {1557-1211},
  month     = jul,
  number    = {1},
  pages     = {347--379},
  volume    = {50},
  doi       = {10.1177/01492063231181651},
  publisher = {SAGE Publications},
}

@Misc{Goel2025,
  author = {Akshay Goel; Atilla Kiraly},
  month  = jul,
  title  = {Introducing LangExtract: A Gemini powered information extraction library},
  year   = {2025},
  url    = {https://developers.googleblog.com/introducing-langextract-a-gemini-powered-information-extraction-library/},
}

@Article{Hogan2012,
  author    = {Hogan, Aidan and Zimmermann, Antoine and Umbrich, Jürgen and Polleres, Axel and Decker, Stefan},
  journal   = {Journal of Web Semantics},
  title     = {Scalable and distributed methods for entity matching, consolidation and disambiguation over linked data corpora},
  year      = {2012},
  issn      = {1570-8268},
  month     = jan,
  pages     = {76--110},
  volume    = {10},
  doi       = {10.1016/j.websem.2011.11.002},
  publisher = {Elsevier BV},
}

@Misc{Huang2024,
  author    = {Huang, Chensen and Zhu, Guibo and Wang, Xuepeng and Luo, Yifei and Ge, Guojing and Chen, Haoran and Yi, Dong and Wang, Jinqiao},
  title     = {Recurrent Context Compression: Efficiently Expanding the Context Window of LLM},
  year      = {2024},
  copyright = {Creative Commons Attribution Non Commercial Share Alike 4.0 International},
  doi       = {10.48550/ARXIV.2406.06110},
  publisher = {arXiv},
}

@InProceedings{Incitti2024,
  author    = {Incitti, Francesca and Salfinger, Andrea and Snidaro, Lauro and Challapalli, Sri},
  booktitle = {2024 27th International Conference on Information Fusion (FUSION)},
  title     = {Leveraging LLMs for Knowledge Engineering from Technical Manuals: A Case Study in the Medical Prosthesis Manufacturing Domain},
  year      = {2024},
  month     = jul,
  pages     = {1--8},
  publisher = {IEEE},
  doi       = {10.23919/fusion59988.2024.10706469},
}

@Book{JeanBaptiste2021,
  author    = {Jean-Baptiste, Lamy},
  publisher = {Apress},
  title     = {Ontologies with Python},
  year      = {2021},
  address   = {Berkeley, CA},
  isbn      = {9781484265512},
  pagetotal = {344},
  ppn_gvk   = {174925462X},
  subtitle  = {Programming OWL 2.0 Ontologies with Python and Owlready2},
}

@Misc{Jiang2019,
  author    = {Jiang, Zhengbao and Xu, Frank F. and Araki, Jun and Neubig, Graham},
  title     = {How Can We Know What Language Models Know?},
  year      = {2019},
  copyright = {arXiv.org perpetual, non-exclusive license},
  doi       = {10.48550/ARXIV.1911.12543},
  publisher = {arXiv},
}

@Article{Ji2022,
  author    = {Ji, Ziwei and Lee, Nayeon and Frieske, Rita and Yu, Tiezheng and Su, Dan and Xu, Yan and Ishii, Etsuko and Bang, Yejin and Chen, Delong and Dai, Wenliang and Chan, Ho Shu and Madotto, Andrea and Fung, Pascale},
  title     = {Survey of Hallucination in Natural Language Generation},
  year      = {2022},
  copyright = {Creative Commons Attribution 4.0 International},
  doi       = {10.48550/ARXIV.2202.03629},
  publisher = {arXiv},
}

@Misc{Jin2024,
  author    = {Jin, Hongye and Han, Xiaotian and Yang, Jingfeng and Jiang, Zhimeng and Liu, Zirui and Chang, Chia-Yuan and Chen, Huiyuan and Hu, Xia},
  title     = {LLM Maybe LongLM: Self-Extend LLM Context Window Without Tuning},
  year      = {2024},
  copyright = {Creative Commons Attribution 4.0 International},
  doi       = {10.48550/ARXIV.2401.01325},
  publisher = {arXiv},
}

@InProceedings{Kim2016,
  author    = {Kim, Jinkyu and Ha, Heonseok and Chun, Byung-Gon and Yoon, Sungroh and Cha, Sang K.},
  booktitle = {2016 IEEE 32nd International Conference on Data Engineering (ICDE)},
  title     = {Collaborative analytics for data silos},
  year      = {2016},
  month     = may,
  pages     = {743--754},
  publisher = {IEEE},
  doi       = {10.1109/icde.2016.7498286},
}

@Misc{Kommineni2024,
  author    = {Kommineni, Vamsi Krishna and König-Ries, Birgitta and Samuel, Sheeba},
  title     = {From human experts to machines: An LLM supported approach to ontology and knowledge graph construction},
  year      = {2024},
  copyright = {Creative Commons Attribution 4.0 International},
  doi       = {10.48550/ARXIV.2403.08345},
  publisher = {arXiv},
}

@Article{LaRocca2009,
  author    = {La Rocca, Gianfranco and van Tooren, Michel J. L.},
  journal   = {Journal of Aircraft},
  title     = {Knowledge-Based Engineering Approach to Support Aircraft Multidisciplinary Design and Optimization},
  year      = {2009},
  issn      = {1533-3868},
  month     = nov,
  number    = {6},
  pages     = {1875--1885},
  volume    = {46},
  doi       = {10.2514/1.39028},
  publisher = {American Institute of Aeronautics and Astronautics (AIAA)},
}

@InProceedings{Madani2019,
  author    = {Madani, Kurosh and Russo, Cristiano and Rinaldi, Antonio M.},
  booktitle = {2019 IEEE International Conference on Big Data (Big Data)},
  title     = {Merging Large Ontologies using BigData GraphDB},
  year      = {2019},
  month     = dec,
  pages     = {2383--2392},
  publisher = {IEEE},
  doi       = {10.1109/bigdata47090.2019.9005991},
}

@TechReport{NASA2017,
  author = {NASA},
  title  = {The NASA Air Traffic Management Ontology},
  year   = {2017},
}

@Misc{Pan2023,
  author    = {Pan, Jeff Z. and Razniewski, Simon and Kalo, Jan-Christoph and Singhania, Sneha and Chen, Jiaoyan and Dietze, Stefan and Jabeen, Hajira and Omeliyanenko, Janna and Zhang, Wen and Lissandrini, Matteo and Biswas, Russa and de Melo, Gerard and Bonifati, Angela and Vakaj, Edlira and Dragoni, Mauro and Graux, Damien},
  title     = {Large Language Models and Knowledge Graphs: Opportunities and Challenges},
  year      = {2023},
  copyright = {Creative Commons Attribution 4.0 International},
  doi       = {10.48550/ARXIV.2308.06374},
  publisher = {arXiv},
}

@Article{Saberi2025,
  author    = {Saberi, Mohammad Ali and Mcheick, Hamid and Adda, Mehdi},
  journal   = {Information},
  title     = {From Data Silos to Health Records Without Borders: A Systematic Survey on Patient-Centered Data Interoperability},
  year      = {2025},
  issn      = {2078-2489},
  month     = feb,
  number    = {2},
  pages     = {106},
  volume    = {16},
  doi       = {https://doi.org/10.3390/info16020106},
  publisher = {MDPI AG},
}

@MastersThesis{Sng2019,
  author = {Sng, Zheng Yang},
  school = {MIT},
  title  = {A Petri Net framework for the representation and analysis of aircraft turnaround operations},
  year   = {2019},
}

@InBook{Stjepandic2015,
  author    = {Stjepandić, Josip and Verhagen, Wim J. C. and Liese, Harald and Bermell-Garcia, Pablo},
  pages     = {255--286},
  publisher = {Springer International Publishing},
  title     = {Knowledge-Based Engineering},
  year      = {2015},
  isbn      = {9783319137766},
  booktitle = {Concurrent Engineering in the 21st Century},
  doi       = {10.1007/978-3-319-13776-6_10},
}

@Article{Villela2005,
  author    = {Villela, Karina and Santos, Gleison and Schnaider, Lílian and Rocha, Ana Regina and Travassos, Guilherme Horta},
  journal   = {Journal of the Brazilian Computer Society},
  title     = {The use of an enterprise ontology to support knowledge management in software development environments},
  year      = {2005},
  issn      = {1678-4804},
  month     = jun,
  number    = {2},
  pages     = {45--59},
  volume    = {11},
  doi       = {10.1007/bf03192375},
  publisher = {Springer Science and Business Media LLC},
}

@inproceedings{wang2024aviationgpt,
  title={AviationGPT: A large language model for the aviation domain},
  author={Wang, Liya and Chou, Jason and Tien, Alex and Zhou, Xin and Baumgartner, Diane},
  booktitle={AIAA AVIATION FORUM AND ASCEND 2024},
  pages={4250},
  year={2024}
}

@Article{Weick1990,
  author    = {Weick, Karl E.},
  journal   = {Journal of Management},
  title     = {The Vulnerable System: An Analysis of the Tenerife Air Disaster},
  year      = {1990},
  issn      = {1557-1211},
  month     = sep,
  number    = {3},
  pages     = {571--593},
  volume    = {16},
  doi       = {10.1177/014920639001600304},
  publisher = {SAGE Publications},
}

@inproceedings{wong2022domain,
  title={Domain Ontology Development Methodology for Construction Contract},
  author={Wong, Saika and Yang, Jianxiong and Zheng, Chunmo and Su, Xing},
  booktitle={International Symposium on Advancement of Construction Management and Real Estate},
  pages={1566--1575},
  year={2022},
  organization={Springer}
}

@Article{Xu2025,
  author    = {Xu, Yiwu and Chen, Yun},
  journal   = {Scientific Reports},
  title     = {Attention-based interactive multi-level feature fusion for named entity recognition},
  year      = {2025},
  issn      = {2045-2322},
  month     = jan,
  number    = {1},
  volume    = {15},
  doi       = {10.1038/s41598-025-86718-0},
  publisher = {Springer Science and Business Media LLC},
}

@Misc{Zhang2024,
  author    = {Zhang, Bowen and Soh, Harold},
  title     = {Extract, Define, Canonicalize: An LLM-based Framework for Knowledge Graph Construction},
  year      = {2024},
  copyright = {Creative Commons Attribution 4.0 International},
  doi       = {10.48550/ARXIV.2404.03868},
  publisher = {arXiv},
}

@article{zhang2026aviationcopilot,
  title={AviationCopilot: Building a reliable LLM-based Aviation Copilot inspired by human pilot training},
  author={Zhang, Zhuorui and Feng, Shanshan and Yang, Tiance and Huang, Ruobing and Wang, Hao and Wang, Fu and Li, Fan},
  journal={Advanced Engineering Informatics},
  volume={69},
  pages={103806},
  year={2026},
  publisher={Elsevier}
}

@Misc{pdf242026,
  author = {pdf24},
  month  = feb,
  title  = {PDF to Text},
  year   = {2026},
  url    = {https://tools.pdf24.org/en/pdf-to-txt},
}
\end{document}